# TRANSFORMER NETWORK FOR MULTI-PERSON TRACKING AND RE-IDENTIFICATION IN UNCONSTRAINED ENVIRONMENT

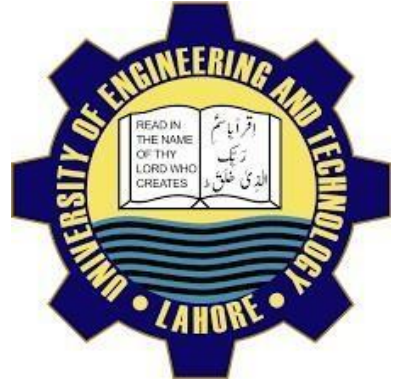

By

## Hamza Mukhtar

hamza.mukhtar@kics.edu.pk

## Research Supervisor:

## Dr. Muhammad Usman Ghani Khan

usman.ghani@uet.edu.pk

## Department of Computer Science
## University of Engineering and Technology, Lahore

.

# ABSTRACT

Multi-object tracking (MOT) has profound applications in a variety of fields including human surveillance, sports analytics, autonomous driving, and cooperative robotics. Despite considerable advancements, existing MOT methodologies tend to falter when faced with non-uniform movements, occlusions, and appearance-reappearance scenarios of the objects. Recognizing this inadequacy, we put forward an integrated MOT method that not only marries object detection and identity linkage within a singular, end-to-end trainable framework but also equips the model with the ability to maintain object identity links over long periods. Our proposed model, named STMMOT, is architectured around 4 key Transformer-based modules: 1) candidate proposal generation, which generates object proposals via a vision-Transformer encoder-decoder architecture that detects the object from each frame in the video; 2) Scale variant pyramid, progressive pyramid structure to learn the self-scale and cross-scale similarities in multi-scale feature maps; 3) Spatio-temporal memory encoder, extracting the essential information from the memory associated with each object under tracking; and 4) Spatio-temporal memory decoder, simultaneously resolving the tasks of object detection and identity association for MOT. Our system leverages a robust spatio-temporal memory module that retains extensive historical observations and effectively encodes them using an attention-based aggregator. The uniqueness of STMMOT resides in representing objects as dynamic query embeddings that are updated continuously, which enables the prediction of object states with an attention mechanism and eradicates the need for post-processing. Experimental results show that STMMOT archives scores of 79.8 and 78.4 for IDF1, 79.3 and 74.1 for MOTA, 73.2 and 69.0 for HOTA, 61.2 and 61.5 for AssA, and maintained an ID switch count of 1529 and 1264 on MOT17 and MOT20, respectively. When evaluated on MOT20, it scored 78.4 in IDF1, 74.1 in MOTA, 69.0 in HOTA, and 61.5 in AssA, and kept the ID switch count to 1264. Compared with the previous best TransMOT, STMMOT achieves around a 4.58% and 4.25% increase in IDF1, and ID switching reduction to 5.79% and 21.05% on MOT17 and MOT20, respectively.

# TABLE OF CONTENTS

# 1. INTRODUCTION

## 1.1. INTRODUCTION

In the field of computer vision, one of the most significant and challenging tasks that consistently draws attention is the tracking and reidentification of multiple objects [3, 13, 16]. This is a key aspect of a variety of different applications, spanning numerous fields such as autonomous vehicles [1], sports analytics [49], and human activity recognition [50]. The primary aim of multi-object tracking (MOT) is to continuously detect and localize multiple targets within a sequence of frames while maintaining the individual identity of each target and predicting their movement trajectories within the visual scene [48]. By doing so, a robust MOT system can monitor and track multiple objects simultaneously, enabling applications like autonomous driving to navigate safely, or for sports analytics to accurately track player movements.

Broadly speaking, MOT techniques can be divided into two main categories: online [13, 16] and offline methods [35]. Offline MOT systems take advantage of the global information available from a complete sequence of frames. This type of method essentially has the advantage of "future" knowledge since it has access to the entire video sequence before making any tracking decisions. It analyses the data from all frames in the sequence to track objects, which enables it to account for events that occur later in the sequence. In contrast, Online MOT operates in a more real-time manner, focusing on identifying a set of objects and establishing their trajectories over frames while ensuring that each object's identification is consistent throughout the sequence. Unlike offline methods, online techniques do not have access to future frames and must make decisions based on current and past frames only. This characteristic makes online methods particularly relevant for real-time applications where immediate decisions must be made, such as in autonomous vehicles or real-time sports analytics. In recent years, online MOT methods have seen a surge in interest due to their practicality and alignment with real-world applications. They are employed in a wide range of scenarios, such as analyzing player movements in football matches [49], tracking the fast-paced action in table tennis games [74], and recognizing human activities in video streams [50]. These real-time applications necessitate reliable and efficient tracking techniques capable of dealing with the complexities inherent in dynamic, ever-changing scenes.

Traditional online multi-object tracking (MOT) methodologies typically hinge on two primary, distinct stages: the object detection phase and the data association phase, also known as object re-identification (ReID). The Object detection stage primarily centers on identifying and pinpointing the exact locations of all desired object instances within each frame of the sequence [16, 2]. It's the process of discovering and accurately locating individual objects within the image data. Sophisticated algorithms sift through the video footage, discerning patterns and features that align with pre-defined models of the objects being tracked. This could be as simple as recognizing basic shapes and sizes, or as complex as identifying specific facial features in a crowd. Each detected object is represented in a form that is conducive to further processing, generally as a set of coordinates that delineate a bounding box around the object in the frame. The Object Re-identification (ReID) stage, also referred to as the data association phase, involves establishing links between detected objects across time [3, 16]. It is in this stage that detected objects from the previous phase are associated with their corresponding instances in subsequent frames, maintaining the continuity of the tracked object's identity over time. The ReID phase essentially simulates the state alterations of the tracked objects, capturing changes in position, size, and other significant attributes, and then faces the challenge of resolving a matching problem between these tracked objects and the outcomes generated from the object detection phase. By successfully associating each object instance with its corresponding presence across frames, the system ensures a coherent, uninterrupted tracking of the object over time. The combination of these two stages - object detection and object re-identification - forms the backbone of traditional online MOT systems. They not only facilitate the tracking of multiple objects in real-time but also ensure continuity and coherence in object identities across multiple frames, thereby significantly enhancing the overall functionality and effectiveness of the MOT system. This dual-phase approach enables the system to accurately track objects in complex, dynamic environments, making it an invaluable tool in a wide range of applications, from autonomous vehicles to video surveillance to sports analytics. The general concept of online MOT is shown in Fig. 1.

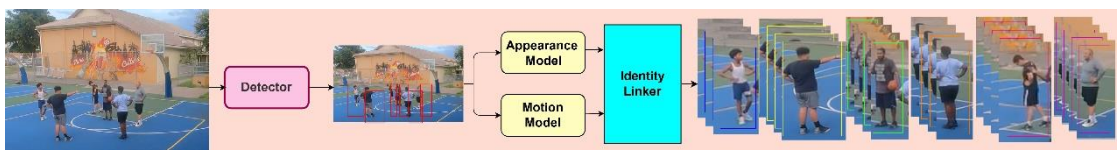

Figure 1.1 Concept of Traditional Online MOT Where Objects are Detected Using An Object Detector and Then Appearance and Motion Models Are Used in Post-Detection Association Linker for the Re-Identification

## 1.2. PERSON TRACKING AND IDENTIFICATION

Multi-Person Tracking and Re-identification [30, 32, 36] constitutes a specialized field of study that aims to address two critical and interconnected challenges: tracking multiple individuals within a dynamic scene and re-identifying these individuals across different time frames or views. Firstly, multi-person tracking involves persistently monitoring the position and movement of multiple individuals within a sequence of video frames. This task requires the capability to handle occlusions, varying individual appearances, and changes in scene dynamics. Not only must the tracking model be able to maintain the identity of each individual over time, but it must also possess the capability to add new individuals entering the scene or remove those who exit. The second component, re-identification, serves to recognize the same individual across different camera views or at different time points. This is a challenging task due to variations in camera views, lighting conditions, individual poses, and occlusions. Re-identification requires learning distinctive and robust appearance features that can help in associating an individual's identity correctly over different spatiotemporal contexts. Thus, the field of Multi-Person Tracking and Re-identification aims to develop innovative algorithms and methods that can seamlessly integrate and perform these two tasks. The resulting models must demonstrate efficiency, robustness, and accuracy in tracking and re-identifying individuals within dynamic and complex environments. These models have vast potential for practical applications in areas such as surveillance, autonomous vehicles, and human activity analysis.

## 1.3.  TRADITIONAL MULTI-OBJECT TRACKING

The methodology behind a majority of online MOT techniques [3, 9, 10] is relatively sequential. Initially, these techniques employ a detector to yield bounding boxes that encapsulate all objects within the current frame. Following this, ReID features are extracted for each bounding box. The next step entails the matching of candidate boxes to existing object trajectories based on the extracted ReID features. However, this workflow comes with its own set of challenges. A prime concern is that the process demands the extraction of ReID features for every box. This necessitates a significant computational load, thereby introducing potential inefficiencies in the system. To mitigate these computational overheads, the field witnessed the advent of the Joint Detection and Embedding (JDE) methods [11, 12, 13, 14, 15, 16]. The innovative approach behind JDE methods lies in their consolidation of the object detection and ReID feature extraction modules into a single network. This unified network has the capability to concurrently predict an object's location and extract its ReID features. By this means, the JDE methods introduce significant efficiencies into the process of online MOT. However, the fusion of object detection and ReID tasks within a single model is not without its complications. There exists a potential for a competitive dynamic to arise between the two tasks, given their disparate nature and requirements. This competitive relationship could inadvertently impact the accuracy of the tracking process, resulting in less reliable results [51-52]. Thus, while the JDE methodology introduces efficiencies, it is not without its trade-offs, necessitating continued exploration and optimization in the field of online MOT.

The objective of the detection task in MOT is inherently geared towards objects within the same category sharing analogous semantics. The goal is to leverage the network's capability to minimize intra-class variation, ultimately refining the detection process. In contrast, the re-identification (ReID) task has a different focus. It centers around the identification of discrepancies among distinct objects falling within the same category. The ReID task, therefore, relies on the network's capacity to amplify and highlight these differences, thereby enabling effective re-identification. This dichotomy presents a predicament for the optimization of the network's performance. The divergent objectives of the two tasks create an incongruity that hinders the network's ability to optimize both tasks simultaneously [29]. This conundrum arises from the fundamentally contradictory goals of the two tasks – one seeking to minimize intra-class variation, the other aiming to accentuate it. Consequently, achieving an optimal balance between these two tasks within the same network remains a complex challenge. Recently, studies have indicated that combining these two stages could yield certain advantages [25, 30]. This integrated approach simplifies the system architecture and potentially enhances computational efficiency. However, this consolidation often inadvertently compromises the association module's ability to model the temporal evolution of objects. In essence, the simplified model may fail to adequately capture the dynamics of object states over time. This can potentially impair the accuracy and reliability of tracking results, necessitating further research to strike a balance between efficiency and performance in MOT systems.

## 1.4. ADVANCE MULTI-OBJECT TRACKING

Recent developments in the field of computer vision have witnessed a growing interest in the application of transformer architectures [18] for a variety of tasks. This includes areas such as object detection [17, 19, 20, 27], person re-identification (Re-ID) [32, 33], semantic segmentation [22], and image super-resolution [21]. These methodologies demonstrate the inherent advantages of attention-based mechanisms, which have been central to the success of transformers in various domains. Transformers possess a unique capacity for concurrently modeling dependencies among multiple input components. They are designed to analyze each input as a part of a comprehensive system, enabling them to make decisions that take into account the entire set of input data. This holistic approach to decision-making is a distinctive feature of transformer architectures, and it is this attribute that enables them to excel in a wide array of computer vision tasks. The benefits of transformer architectures are particularly pertinent to the challenges faced in MOT. One of the main challenges in MOT is to accurately model the interactions between different objects, especially in scenes that are densely populated with numerous objects. Given the complexity and dynamism of these scenes, accurately predicting the state of each object and their interactions is a non-trivial task. The strong ability of transformers to model relationships between various entities is especially suited to address this challenge. By incorporating transformers into MOT systems, it is feasible to improve the accuracy of object interaction modeling, and subsequently, enhance the overall tracking performance. This underscores the potential of transformers as an effective solution for the intricate challenges posed by MOT tasks.

## 1.5. RESEARCH CONTRIBUTION

This works presents the develops a novel Transformer-based model for MOT and ReID, referred to as SpatioTemporal Multi-Object Tracking (STMMOT). Unlike traditional models, STMMOT executes object detection and identity association in tandem within a singular framework, while maintaining an online operating mode. This unified approach propels efficiencies in processing and promises to improve the accuracy of outcomes. STMMOT is underpinned by a key feature – the employment of an expansive spatio-temporal memory. This memory apparatus records prior observations of tracked objects and serves as a dynamic reference reservoir for the model. At each time step, the memory undergoes active encoding by assimilating and reflecting relevant information from recent observations. This dynamic encoding allows the system to approximate object states with higher precision, a critical function in successful identity association tasks. The spatio-temporal memory contributes towards a comprehensive depiction of tracked objects. This rich representation becomes instrumental when resolving object detection and association tasks within a unified decoding module, a central component of STMMOT. This module is responsible for producing direct outputs of tracked object instances, which include both recurring and newly encountered objects within the latest frames. The theoretical construct of STMMOT is represented pictorially in Figure 2. This figure offers a visual understanding of the key components and their interplay within the system, contributing to the holistic understanding of the developed model. Our contribution through STMMOT advances the state-of-the-art in Transformer-based MOT models, showcasing the practical potential of this novel approach in complex tracking scenarios.

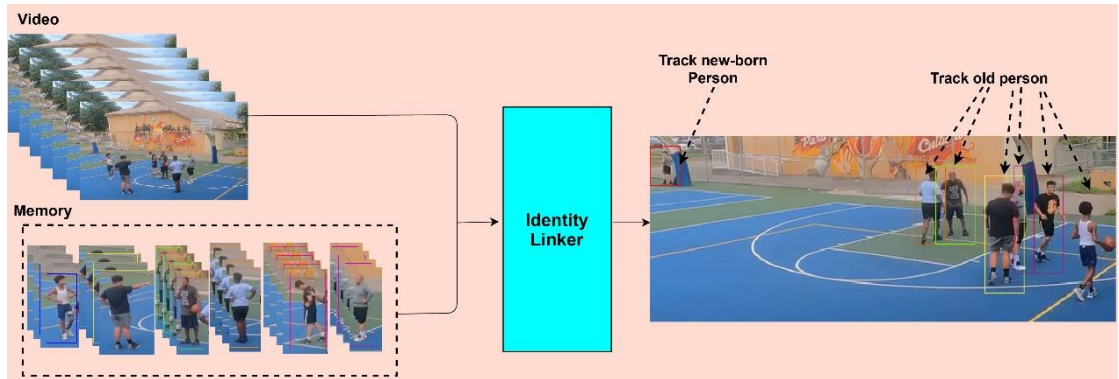

Figure 1.2 The Concept of STMMOT is Depicted Through the Use of a Dynamic Spatio-Temporal Memory Buffer that Keeps the Object States of All Tracked Objects and Updates Them Over Time.

The main contributions for face reconstruction are eew:

1.  The development of a new end-to-end MOT model, named SpatioTemporal Multi-Object Tracking (STMMOT), is introduced. This novel system simultaneously learns and applies object detection and identity association. It merges these two traditionally separate tasks into a unified network, marking a shift from the conventional models and paving the way for more efficient and effective tracking strategies.

2.  A key feature of STMMOT is the integration of spatiotemporal memory. This innovative component actively models both temporal and spatial contexts of target objects. By doing so, it facilitates robust object association over time and enhances target-specific discriminative capabilities. The incorporation of such a mechanism is a significant step forward, ensuring robust MOT, even in complex and dynamic environments.

3.  In addition to the spatiotemporal memory, we propose a multi-scale attention pyramid module. This module efficiently leverages cross-scale similarities of the same object tracklets across multiple time points. It ensures a comprehensive utilization of the available information, thereby enhancing the accuracy and reliability of the tracking outcomes.

4.   Lastly, the performance of our approach is validated through a series of comprehensive experiments conducted on popular MOT datasets, including MOT17 [54] and MOT20 [55]. These experimental results demonstrate that our approach significantly improves the association capabilities of re-identification (ReID) features. The performance metrics showcase that our STMMOT model is highly competitive and offers substantial improvements compared to existing mainstream algorithms in the field of MOT.

## 1.6. THESIS ORGANIZATION

The rest of the text is sectioned as follows. Chapter 2 briefly describes the traditional object tracking and reidentification methods and then switches focus to more advance MOTs. This chapter aims to describe the different types of previous MOT methods such as tracking-by-detection, joint-detection-tracking, transformer--tracking and memory-tracking, and identify the research gap. Chapter 3 explains the research gap in the previous work, while. Chapter 4 discusses the problem statement on which this study focuses. Chapter 5 describes the experimental setup used while evaluating the aforementioned methods. Chapter 6 presents an evaluation of the effectiveness and efficiency of our methodology through various ablation experiments and then compares it with various state-of-the-art methods. on both public datasets and sports data, and discusses the results. Chapter 7 discusses the conclusion and Future Work.

# 2.   LITERATURE SURVEY

In this chapter, the relevant theory and related work are presented about MOT from traditional to advance methods. Proposed works are separated into categories, namely: (1) tracking-by-detection, (2) joint-detection-and-tracking, (3) transformer-based-tracking, and (4) memory-based-tracking. Each of these categories encapsulates a distinct approach towards the objective of MOT and demonstrates unique attributes that are indicative of their application context.

## 2.1.   TRACKING-BY-DETECTION

As the field of object detection has witnessed significant advancements [1, 2], various multi-object tracking (MOT) methodologies [3, 4] have increasingly adopted the tracking-by-detection strategy. These approaches principally focus on the facet of tracking, that is, establishing consistent identification of the same objects across sequential frames, capitalizing on high-performing detectors that furnish the bounding boxes demarcating the objects. Early research in this domain relied heavily on the motion attributes of objects to predict their spatial locations in forthcoming frames. Techniques such as the Kalman Filter [5] and particle filters [6] were commonly employed for this purpose. A notable pioneer, the Sort model [4], utilized the Kalman Filter to predict the spatial coordinates of all potential bounding boxes in the subsequent frames. Following this prediction, the Intersection over Union (IOU) between the previously established bounding box and the predicted boxes in the succeeding frame was computed. The Hungarian algorithm was then invoked to match these bounding boxes, thereby establishing the tracking component. Seeking to further refine the tracking-by-detection paradigm, StrongSort [3] introduced an enhanced iteration. This model incorporated the visual appearance information of the objects under tracking into the process. It accomplished this by extracting the visual features of the objects by feeding the cropped region-of-interest pertaining to the object into a separate convolutional neural network (CNN). This integration of appearance data further strengthened the tracking accuracy and robustness. Taking this a step further, [7] integrated human pose information into the tracking process, adding another dimension to the tracking mechanism, particularly valuable for human tracking scenarios. This trend exemplifies how the tracking-by-detection paradigm has been progressively enriched by integrating supplementary sources of data and leveraging advancements in related fields.

The LSTM [8] is employed to predict the object location in the current frame using prior frame data. POI [9] determines the affinity value to assign identify to a detected object by measuring the distance between the visual feature maps of the detected objects. STRN [10] learns to identify similarities between tracked objects by encoding various cues such as visual, motion, and location over a long span through maintaining a spatial-temporal relationship. While the tracking-by-detection pipeline delivers remarkable performance, its model complexity and computational demands are suboptimal. Ideal filters in these methods maintain tracking states with historical information for new frame predictions. Although optimal states can be determined in linear-Gaussian cases, estimating them in non-linear and non-Gaussian scenarios, such as occlusions in visual multi-object tracking, is challenging due to finite-dimensional state representation cons.

## 2.2. JOINT-DETECTION-TRACKING

Joint object detection and tracking pipelines aim to concurrently execute the tasks of detection and tracking within a single-stage process. These methods represent an effort to streamline the process and reduce computational overhead by fusing the traditionally separate stages of object detection and tracking. A notable instance of this methodology is the D&T [11] model, which presents an end-to-end spatio-temporal multi-object tracking approach that is jointly trained. Unique in its design, it adopts a siamese architecture that concurrently tackles two tasks: object detection in the current frame and capturing cross-frame co-occurrences for object tracking. This dual-focus mechanism allows for real-time tracking of objects as they traverse through the sequence of frames. Building on this concept, Tracktor [13] has introduced an innovation where the tracking boxes from the previous frame are utilized for regression. This allows the model to establish a continuity of object presence across frames, thus enhancing tracking accuracy. Integrated-Detection [12] further refines this joint detection and tracking approach. It enhances the detection output by integrating detection boxes within the region proposals. It then applies box offset regression to the combined proposals to compute tracking coordinates in the current frame. In effect, this innovative strategy eliminates the need for a separate identity association stage, thereby further streamlining the tracking process. These advancements illustrate the potential for integrating object detection and tracking into a unified pipeline, demonstrating how combining these tasks can increase efficiency while maintaining, or even enhancing, tracking accuracy.

Centertrack [14] leverages prior frame feature embeddings as supplementary input for estimating the motion direction of the object's central point between the part and current frames. Despite the incorporation of detection and tracking details within a unified network, effectively dealing with intricate scenarios continues to pose a significant challenge. Consequently, the JDE [15] suggests incorporating an embedding layer to the detector for ReID feature extraction and appearance integration where detection and identity association share same features to enables the model to produce both detection outcomes and associated ReID features jointly, thus enhancing MOT accuracy and speed. FairMOT [16], a multi-task MOT network, detects the object and extracts visual embedding from a shared backbone to improve association accuracy. While this joint end-to-end MOT tries an equilibrium between accuracy and efficiency, the competition between object detection and ReID persists, negatively impacting network performance In summary, these methods reduce the computation but compromise tracking recovery after occlusion and are unable to re-establish connections with objects missing for extended periods.

## 2.3. TRANSFORMER-TRACKING

The transformer architecture [17, 21] has demonstrated its potency and impact on vision tasks. As a unique query-key mechanism, the transformer primarily depends on the attention mechanism to process the deep features extracted. Initially exhibiting exceptional efficiency in natural language processing [18], it later transitioned to visual perception tasks [19], achieving notable success. The transformer's elegant structure and impressive performance have attracted the vision community. It has exhibited significant potential in detection [20, 19, 21], segmentation [22], and 3D data processing [23]. Transformers have only recently been employed in MOT, before transformers, various attention mechanisms were introduced for MOT applications. Specifically, [24] suggests a target-aware and distractor-aware attention mechanism to generate more dependable visual embeddings, which also aids in suppressing old objects. Following the success of transformers in detection, two concurrent works, TrackFormer [25] and MOTR [26], apply vision-transformers based on the DETR framework [27] for the MOT task. Conversely, TransCenter [28] and TransTrack [29] utilize Transformers solely as feature extractors and recurrently pass track object features to learn each object's joint embedding aggressively.

For occluded Re-ID, a dual-branch Transformer [30] has developed comprising a global branch for global feature extraction and a local branch featuring Selective Token Attention for local feature extraction using multi-headed self-attention. Another dual-branch MOT JDE[31] includes a Patch-Expanding mechanism to boost object detection and identify associations in crowded scenarios by enhancing feature maps spatially using CNN and Einops Notation rearrangement. These dual-branch design lacks end-to-end learning and the performance of identity association depends on the first branch. Conversely, an end-to-end transformer-based occluded person Re-ID model [32] leverages a multi-headed self-attention to learn the distribution of common non-occluded target person regions and generates accurate crops using the Minimized Character-box Proposal method. TranSG [33] utilizes a structure-trajectory mechanism to capture both relations and critical spatial-temporal semantics from skeleton graphs to get the fine-grained representations of body joints. The P3AFormer [34] utilizes flow information to guide the propagation of pixel-wise features, generate multi-scale feature embeddings through a meta-architecture, and pixel-wise identity assignment is used to associate connections between the same object appearances in a sequence of frames. Another approach for unsupervised transformer Re-ID [35] is proposed where local tokens are divided into multiple parts to create the part-level feature embeddings, while global tokens are averaged to produce a global embedding. The DC-Former [36] further improves the representation discrimination by dividing the embedding into multiple diverse and compact subspaces. TransMOT [37] employs a graphical transformer to model spatio-temporal interactions between objects by representing object trajectories as sparse weighted graphs. Although the aforementioned work focuses on representing object states using dynamic embeddings, it lacks adequate modeling of long-term spatiotemporal information and adaptive feature fusion methods.

## 2.4. MEMORY-TRACKING

Memory networks, a form of recurrent neural network (RNN) that can maintain a long-term internal state or 'memory' over time, have gained widespread acceptance in various fields of machine learning. They have demonstrated significant utility in natural language processing (NLP), where they assist in understanding long-term dependencies and temporal reasoning in tasks such as textual question-answering [40] and task-oriented dialogue systems [41]. In addition to NLP, they have been effectively utilized in spatio-temporal video analysis [42] as well, where they store and retrieve time-indexed features to accurately recall historical scene information over an extended duration. More recently, the use of memory networks has been explored in the domain of multi-object tracking (MOT), which has a crucial requirement for both spatial and temporal information. Notable examples of this trend include MemTrack [43] and STMTrack [44]. However, these initial forays into the utilization of memory networks in MOT have largely been confined to acquiring object appearance information. They have largely overlooked the inherent relationships between memory frames and the embedded temporal context, which are critical elements for improving tracking accuracy over time. A more comprehensive utilization of memory networks can significantly enhance their potential in MOT applications. In particular, robust object association across time - a key challenge in MOT - calls for the implementation of large spatiotemporal memory. The memory network, by encoding and recalling spatial and temporal cues about the objects, can thereby contribute significantly to improving tracking accuracy and robustness. It is thus imperative for future MOT approaches to fully exploit the potential of memory networks in addressing the complex spatio-temporal challenges inherent to multi-object tracking.

# 3.   RESEARCH GAP

Existing state-of-the-art multi-object tracking models primarily rely on Intersection-over-Union (IoU), velocity, and a fixed-size feature vector of the bounding box, exhibiting shortcomings in various real-world scenarios. They grapple with dynamic changes in object velocity and trajectory, leading to inconsistencies in tracking performance. Challenges also surface when dealing with intersections or overlapping objects and temporary or prolonged disappearances and reappearances of objects, often resulting in identity switches. Occlusions, or cases where an object is partially or fully blocked from view, can disrupt the object detection process, causing tracking inaccuracies. Additionally, these models struggle to maintain consistent performance when faced with significant variations in video or object resolution. They also lack robustness against variations in the scale of objects within a frame and tend to falter when confronted with changes in object speed or angular movement. This underscores a substantial research gap in improving the adaptability and robustness of tracking models to handle these varied conditions, underlining the need for more accurate and efficient multi-object tracking systems.

# 4.    PROBLEM STATEMENT

Existing state-of-the-art systems for person tracking typically assume consistent object motion and minimal occlusion, limiting their efficacy under complex, real-world scenarios. When confronted with variable object speeds or situations where the target is obscured by other individuals or objects, their performance markedly deteriorates. This presents a considerable challenge in environments with unconstrained movement and high degrees of occlusion, where maintaining accurate, continuous tracking is paramount. In response to this problem, our research proposes a novel, end-to-end learnable model, designed to concurrently execute object detection and identity association tasks. Central to our approach is the introduction of a spatiotemporal memory buffer, a unique structure that continuously stores and updates information about each tracked object. This buffer provides a dynamic record of all objects in the scene and is critical to maintaining accurate object identities, even under challenging conditions such as occlusions or rapid changes in object motion. This approach offers a significant advancement in multi-person tracking methodologies, facilitating the handling of complex scenarios in unconstrained environments. By integrating detection and identity association within a unified framework and leveraging a spatiotemporal memory buffer, we aim to significantly enhance the robustness and accuracy of multi-person tracking systems.

# 5.   METHODOLOGY

The ideal outcome for MOT is a complete and accurately ordered set of objects for every frame within a video. Given a series of frame sequences denoted as $I = \{I_0, I_1, I_2, \ldots, I_T\}$, MOT aims to identify and track $N$ object's locations while maintaining their trajectories $T = \{T_0, T_1, T_2, \ldots, T_N\}$ through real-time processing. we introduce STMMOT which simultaneously learns object detection and identity association. Unlike proposed MOT techniques [3, 34, 28, 31] that only transfer tracked object states between consecutive frames, our method incorporates a spatio-temporal memory block for storing long-term dependencies of tracking objects. Additionally, we employ a memory encoder-decoder mechanism that effectively extracts relevant representation for associating the same objects in a sequence of frames even after an extended period.

## 5.1.   END-TO-END MECHANISM

As illustrated in Figure 3, STMMOT is comprised of four primary modules: 1) a frame-level candidate proposal generation module $\Theta_C$, which detects objects and creates the object proposals against each frame for the current time step $I_t$, 2) a scale variant progressive pyramid module $\Theta_P$ to learn the cross-scale similarities for handling the varying sizes and scales within a frame sequence, 3) a track-level memory encoding module $\Theta_{ME}$, responsible for aggregating associated object embeddings, and 3) a memory decoder $\Theta_{MD}$, which associates detected proposal object candidates with already tracked objects. At time step t, $\Theta_C$ creates $N^t{}_{can}$ candidate object proposals denoted as proposal embeddings $Q^t{}_{can} \in R^{N^t{}_{can} \times d}$, using a Transformer encoder-decoder network. $\Theta_{ME}$ forms the signal compact representation from the historical states of each tricked object dynamically, referred to as tracklet embeddings $Q^t{}_{traklet} \in R^{N^t{}_{tracklet} \times d}$. Taking encoded feature map of the object as query with $[Q^t{}_{can},$ $Q^t{}_{tracklet}$], $\Theta_{MD}$ determines and measures the inter-object linkage, and updates the embeddings as [$'Q^t{}_{can}$, $'Q^t{}_{tracklet}$]. The locations [$L^t{}_{can}$, $L^t{}_{tracklet}$] and confidence scores [$C^t{}_{can}$, $C^t{}_{tracklet}$] of new and tracked objects are then estimated based on these resultant embeddings. The location vector represents the spatial position of each object in the frame sequence, while the confidence scores represent the model's certainty regarding the presence of those objects. Lastly, the locations and states of already tracked objects are utilized to update their trajectory and memory. Trajectory refers to the path that the object has followed over time, while the memory buffer encapsulates the object's historical states, including its past locations, sizes, and appearance features. The historical data assists in maintaining consistency in object tracking, especially in challenging scenarios like occlusions or the temporary disappearance of the object from the frame. Newly detected objects are added in $T$, and their initial state is stored in the memory buffer. This action prepares the system to monitor this object in subsequent frames, effectively expanding the model's tracking scope.

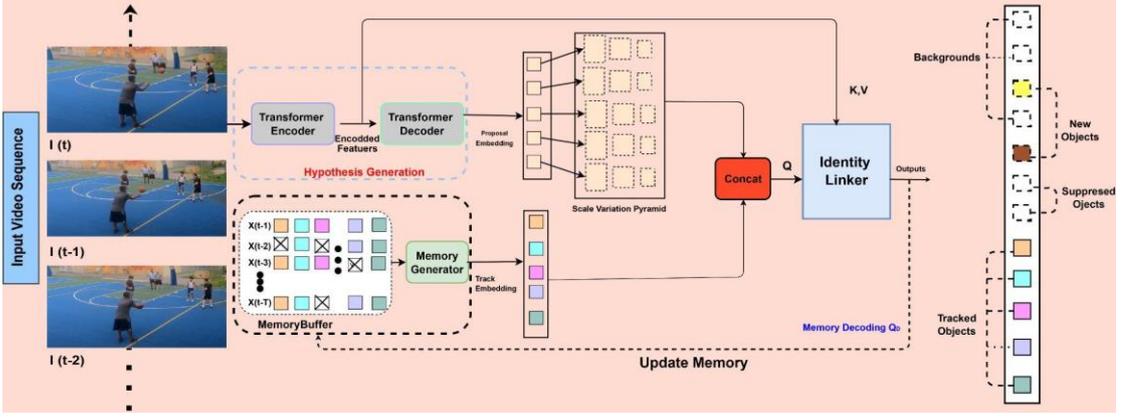

Figure 5.1 STMMOT System Operates with four Primary Units: 1) a candidate proposal creation network $\theta_C$ that generates object proposals for the present frame, 2) Scale Variant Progressive Pyramid module $\theta_P$ that constructs a hierarchical feature pyramid with different levels of resolution to enable the extraction of features from tracklet candidates at multiple and cross scales, 3) a memory encoder $\theta_{ME}$ that extracts essential representation associated with each object under tracking and 4) a memory decoder module $\theta_{MD}$ that concurrently addresses object localization and identity linkage. STMMOT preserves a memory buffer that holds long-term object states, with an encoding-decoding mechanism helping to associate objects over time. Each candidate proposal object is classified as new, tracked, or part of the background.

## 5.2. CANDIDATE PROPOSAL NETWORK

The candidate proposal network (CPN) $\theta_C$ utilizes a vision Transformer encoder-decoder design [26, 27] and a CNN as a feature extractor to create $N^t{}_{can}$ object proposals. These proposals either initiate the tracking of new objects in the current frame or update already tracked objects by providing new spatial and identity information. The $\theta_C$ transformer encoder processes a sequentialized feature map $f^t{}_0$ $\in R^{D \times HW}$, derived from the current step input frame $I_t$ using a CNN backbone where $I_t \in R^{H \times W \times C}$ is fed to a CNN that produces an information-rich but low-resolution representation. This sequential processing enables the model to account for temporal changes in object positions, shapes, and appearance attributes, which is crucial for tracking objects over consecutive video frames. The Encoder network expects to receive a sequence as input, so, spatial dimensions of the feature maps $f^t{}_0$ are transposed into one dimension to form the $D \times HW$ shape feature map. The image features is encoded as $f^t_1 \in R^{d \times HW}$ using stacked Transformer encoder layers. The decoder $\theta_C$ takes the encoded image feature map $f^t_1$ and empty object queries (portrayed as learnable embeddings) and generate the final proposal embeddings $Q^t_{can} \in R^{N^t_{tracklet} \times d}$. Each decoder layer consists of multi-head self-attention (MSA) and a feed-forward network (FFN). As the transformer is permutation-invariant, Each element in $f^t_0$ is enhanced with a distinct positional encoding that is included in each MSA layer to signify its spatial location. The objectness scores and bounding box coordinates for each proposal are predicted using $Q^t_{can}$. Our decoder follows the transformer [27], transforming N embeddings using self-attention and cross-attention. Unlike the transformer [26], our model takes a different approach by simultaneously decoding N objects at each decoder layer. To ensure unique outcomes, the N input embeddings, referred to as object queries, need to be distinct. These object queries are positional encodings that are learned and added to each attention layer's input, similar to the encoder. The decoder transforms the object queries into output embeddings, which are independently decoded into box coordinates and class labels using a feed-forward network (FFN). This process generates N's final predictions. By employing self-attention and cross-attention attention on these embeddings, our model reasons globally about all objects, considering pairwise relations between them. It can also utilize the entire frame as context. The architecture of the CPN is illustrated in Fig. 4.

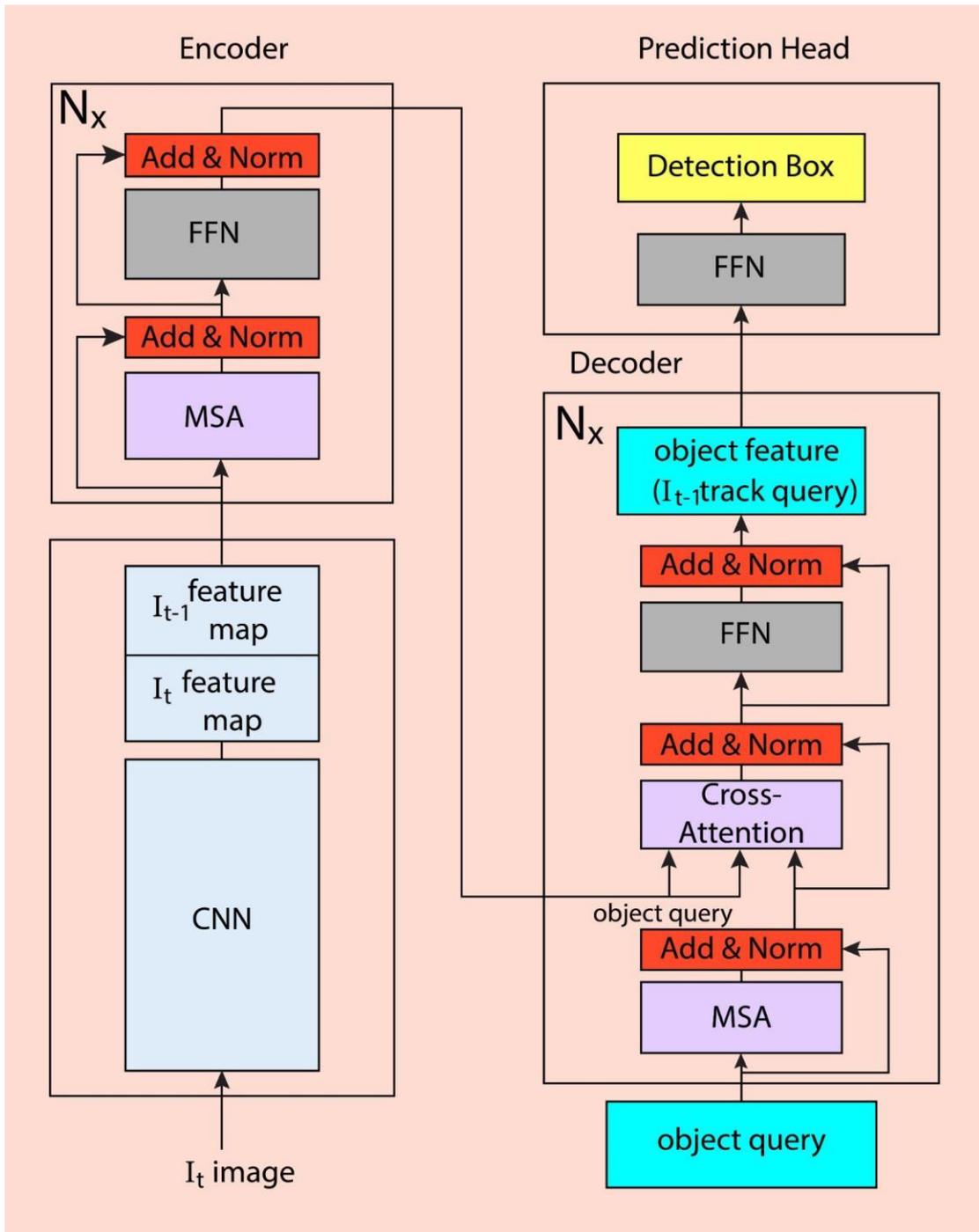

Figure 5.2 The Architecture of the Candidate Proposal Network Based on Vision-Transformer

## 5.3. SCALE VARIANT PYRAMID

The proposed Scale variant pyramid (SVP) detailed architecture is illustrated in Fig. 5. The SVP module accepts input features $X_{[0:N]}$ and requires $N + 1$ executions. It uses a progressive structure to learn similarities between self-scale and multi-scale feature maps. In the pyramid structure, strided convolutions are used to downscale the feature maps from the upper layer by a factor of 2 to obtain feature maps at the current layer. Let M be the pyramid layer in the SVP module, and $N$ be the number of progressive feature transfer layers (PFTLs) at each layer to progressively extract self-similarity. At the m-th layer, the n-th PFTL takes the first feature $X^m{}_0$ as input and the output of the previous block $(X^{n-1}{}_i)^m$, (i $\in$ [0, N]). For the first PFTL, the inputs are $X^m{}_0$ and $X^m{}_i$. Taking inspiration from TDAN [71] and DCN [72], deformable convolution is applied in the PFTL. This process can be expressed as:

$$(X^D{}_i)^{m,n} = X_{Dconv}(X^m{}_0, (X^{n-1}{}_i)^m) \qquad (1)$$

where $X_{Dconv}(\cdot)$ denotes deformable convolution, and $(X^D{}_i)^{m,n}$ represents the output of the n-th block's deformable convolution at the m-th layer. Deformable convolution helps in capturing more complex patterns by introducing learnable offsets to the convolutional kernel, which adds an extra level of adaptability in handling geometric transformations. These learned offsets are predicted based on the input data, acting as an additional layer of complexity to the convolution operation. The advantage of this approach is that it allows for non-linear geometric transformations, providing more robustness to changes in object size, shape, and orientation. The learned offsets can adapt to the specificities of the input, focusing on areas of the image that contain the most relevant information, making the feature extraction process more efficient and accurate.

$$(\Delta P_i)^{m,n} = X_C(X^m{}_0 || (X^{n-1}{}_i)^m) \qquad (2)$$

where $(\Delta P_i)^{m,n}$ denotes the learned offset of the n-th block at the m-th layer, $||$ represents channel-wise concatenation, and $X_C(\cdot)$ indicates the convolution operation.

Learning the offset is a vital step as it allows for subtle adjustments of the feature map locations to better accommodate the object's characteristics. Channel-wise concatenation combines feature maps from different layers so that the subsequent layers can use the combined features. Next, we compute the feature-level mask of the n-th block at the m-th layer $(Mask_i)^{m,n}$ which forces the PFTL to concentrate on the most correlated feature information:

$$(Mask_i)^{m,n} = Softmax(X_C(X^m{}_0) - X_C((X^{n-1}{}_i)^m))$$  (3)

The motion attention mask is then element-wise multiplied with the output of the deformable convolution. This product is then sent through another convolutional layer, yielding a feature set that encapsulates the residual information within this block. The n-th PFTL's output feature at the m-th layer is extracted by adding the residual information to the first feature:

$$(X^n{}_i)^m = X^m{}_0 + X_C \ (X^m{}_0 \ || \ (Mask_i)^{m,n} \ \otimes \ (X^D{}_i)^{m,n})$$  (4)

where $\otimes$ denotes element-wise multiplication. Lastly, the SVP module's output feature at the m-th layer $(X^{SVP}{}_i)^m$ is defined as:

$$(X^{SVP}{}_i)^m = F_C(Us(((X^{SVP}{}_i)^{m+1})^\varepsilon \ || \ (X^N{}_i)^m \ )$$  (5)

where $(X^N{}_i)^m$ denotes the feature generated after the N block of PFTL at the m-th layer. Us($\cdot$) is the scaling of the feature map by a hyperparameter $\varepsilon$, and bilinear interpolation is used for the resizing. The proposed SVP builds a three-layer pyramidical architecture (M = 3) and leans multi-scale dependencies progressively coarse-to-fine fashion. The architecture of the PFTL is shown in Fig. 6.

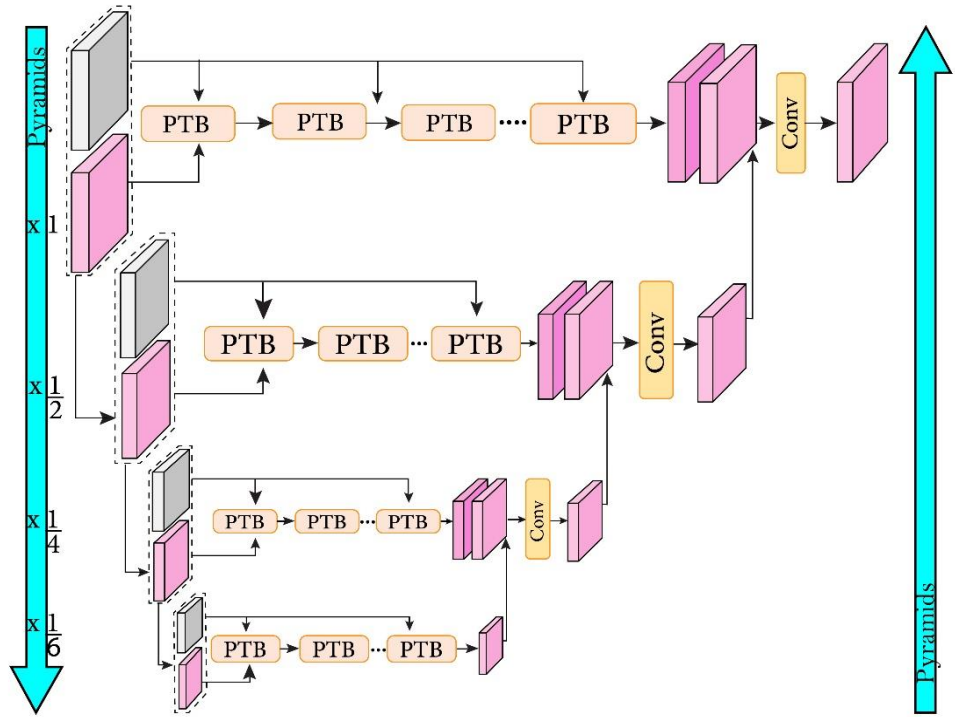

Figure 5.3 Architecture of Scale Varianid Pyranid

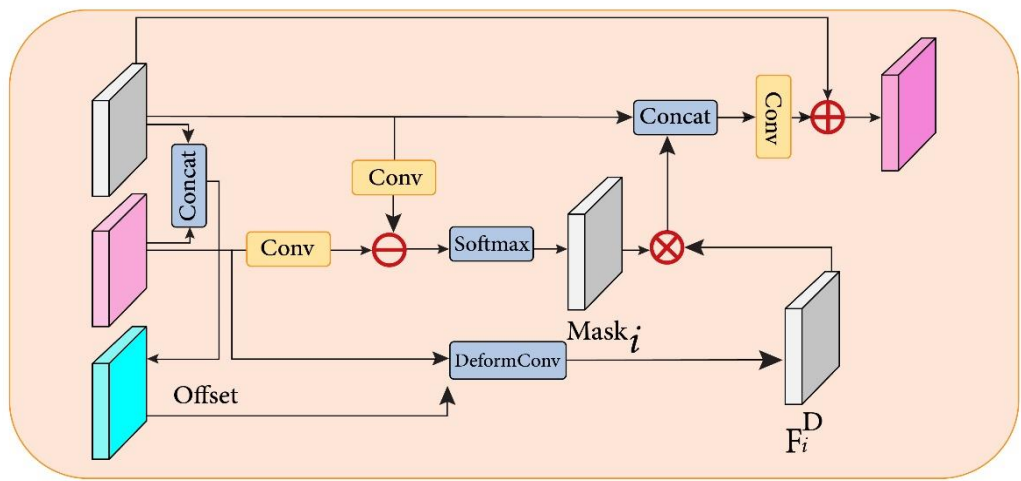

⊕ Element-wise addition ⊖ Element-wise Subtraction ⊗ Element-wise Multiplication

Figure 5.4 Architecture of Progressive Feature Transfer Layer

### 5.4. SPATIO-TEMPORAL MEMORY

MOT requires learning long-term spatio-temporal dependencies in occluded and non-uniformed motion scenarios. This involves preserving the visual-motion features while capturing cross-frame similarities that co-occur both in the visual and spatial domains, which are crucial for effective tracking. Thus, we create a spatio-temporal memory buffer $B \in R^{N \times T \times d}$ to store all N-tracked objects' historical states. The buffer is a three-dimensional structure that can hold N objects (each having a unique history) tracked over T time steps, each having a state in d-dimensions. It captures the spatial location and temporal progression of each object, i.e., where the object is and how it moves over time. This buffer keeps a maximum of $N_{max}$ objects and up to $T_{max}$ time steps for each tracked object. The structure of this memory buffer is built using a first-in-first-out (FIFO) data structure which ensures that the memory buffer does not exceed its capacity and always contains the most recent information. So, the oldest tracked object is first to be suppressed when the buffer reaches its maximum capacity. The maximum capacity is set by two parameters: $N_{max}$ (the maximum number of tracked objects) and $T_{max}$ (the maximum number of time steps for each tracked object). At time step t, the memory represents the states of $N^{t-1}{}_{tracklet}$ active objects in the previous $T$ frames, denoted as:

$$X^{t-1-T:t-1} = \{x_n{}^{t-1-T:t-1}\}_{k=1:N^{t-1}{}_{tracklet}} \tag{6}$$

where $x_n{}^{t-1-T:t-1}$ represents the n-th object's states and its state is padded with 0s if the object is not present in the frame $I_t$. Once the T exceeds $T_{max}$, the initial state $x_n{}^{t-1-T}$ of each tracked object is removed from the memory. The values of $N_{max}$ and $T_{max}$ are chosen based on the specifics of the application and hardware limitations. $N_{max}$ should be large enough to handle the desired number of objects in a frame sequence, e.g., 200 or 400, while $T_{max}$ needs to be large enough that is reasonable for handling the occlusion and encompass the length of time step (e.g. frame sequence) the objects need to be tracked, such as 25 or 50 frames. To remain within the hardware limitation, we set the $T_{max}$ is 30 frames.

## 5.5. MEMORY ENCODING

Fig. 7 illustrates the encoding of candidate memory and extraction of track embedding via three attention blocks. Firstly, a short-term context block, $b_s$, amalgamates embeddings from consecutive frames, aiming to reduce noise. Secondly, a long-term context block, $b_l$, is utilized to derive pertinent features within the timespan encompassed by the memory. Lastly, a fusion block, $b_f$, integrates the embeddings generated from both the short and long-term branches. For the short-term block, the most recent state is used as the query input for the cross-attention mechanism, ensuring the model responds quickly to recent changes in the data. For the long-term block, the input query is an updated embedding called Dynamic Memory Aggregation Tokens (DMAT), which represents an aggregated view of the memory, updated at every timestep. For each tracklet, the short-term block $b_s$ processes it's historical $T_s$ states, while the long-term block $b_l$ employs a comparatively long history length $T_1$ ($T_s <<$ $T_l$). Multi-head cross-attention is used in both short-term context $b_s$ and long-term context $b_l$, with history states serving as key and value inputs, allowing the model to focus on various temporal parts of the states simultaneously. The input query for $b_s$ is the current state $S^{t-1}$, while a dynamically updated embedding, DMAT, $Q^{t-1}{}_{DMAT} = \{q^{t-1}{}_k\}k = 1 : N_{tracklet}$, is utilized for $b_l$.

At the onset of every tracklet, which represents a short sequence of object states, DMAT is allocated. Initially, all tracklets have identical DMATs, symbolizing the same initial memory representation for each tracked object. Afterward, at time step t > 0, the DMATs undergo iterative updates based on the states from the previous time step. This implies that at each step, the DMATs assimilate information from the previous state, updating their values to represent the evolution of the object states over time. This design concept is examined in greater detail and validated in Section 4.4. The outputs from both the short-term and long-term blocks, referred to as Aggregated Short-term Context (ASC) $Q^t_{ASC}$ and Aggregated Long-term Context (ALC) $Q^t_{ALC}$ respectively, are then merged by a fusion block $b_f$. Fusion block combines the information from the short-term and long-term contexts into an aggregated representation $Q^t_{DMAT}$. and produces a consolidated track embedding $Q^t_{tracklet}$. In addition to the track embedding, the fusion block also outputs an updated $Q^t_{DMAT}$. This updated token is saved and used in the subsequent time step, allowing the system to keep track of the ongoing evolution of object states over time. In this way, the model retains and exploits temporal context in tracking objects, enhancing the overall tracking performance.

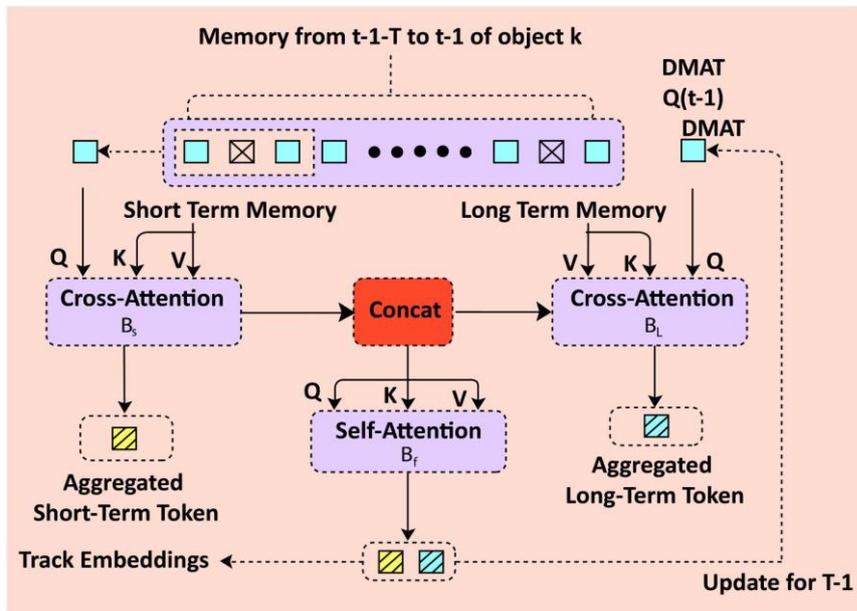

Figure 5.5 The Architecture of Memory Encoder Consists of Three Blocks : short-term memory block ($b_s$) that reduces recent frame noises, long-term ($b_l$) that pulls a larger historical object state context capture more a broad patterns, and a fusion block that combines the states from the short and long term contexts into a single and compact

representation. The merged embeddings are then used as track embeddings (blue-white query) and updated in the DMAT (blue-red query) for the subsequent time step.

## 5.6. MEMORY DECODING

The memory decoder $\theta_{MD}$ is the decision head of the STMOT responsible to output the object identity. It receives the encoded frame feature from the CPN encoder, proposal embedding from the CPN decoder, and track embedding from the memory aggregator to final tracking outcomes. This is achieved through a stack of Transformer decoder modules, using the fused candidate proposal and track embeddings $[Q^t{}_{can},\ Q^t{}_{tracklet}]$ as queries. $\theta_{MD}$ utilizes the encoded frame feature $x^t{}_1$ from $\theta_C$ as key-value pair. In $\theta_{MD}$ output $['Q^t{}_{can}, 'Q^t{}_{tracklet}]$, each entry $q^t{}_i$ goes through a decoding process that transformer the fused embedding into three predictions: bounding box coordinates, objectness, and uniqueness score. Through the use of the Transformer, decoder applied to these embeddings, the model reasons globally about all the candidate proposals via pair-wise relation, while concurrently incorporating the full frame as a contextual reference. The objectness score $o^t{}_i$ and uniqueness score $u^t{}_i$ ranges from 0 to 1, where $o^t{}_i = 1$ indicates that the model identifies a visible object and $u^t{}_i = 1$ predicts that object is unique and needs to be included in the tracking results. Otherwise, it needs to be suppressed. We assume $u^t{}_i = 1$ when $q^t{}_i \in\ 'Q^t{}_{tracklet}$. When a proposal is determined to be unconnected to any identity currently being tracked, it is regarded to be novel ($u^t{}_i = 1$), and is assigned a uniqueness value $u^t{}_i$ of 1. Consequently, a unified confidence score, which applies to proposal and track entries, is established. This score is calculated as the product of objectiveness and uniqueness scores.

$$s^t{}_i = o^t{}_i . u^t{}_i \qquad (2)$$

The model predicts two types of confidence score predictions for the tracking process, $s^t{}_{can}$ for candidate proposal and $s^t{}_{tracklet}$ for track queries. The confidence scores, calculated by the model, evaluate the probability of a proposal or track query representing a valid object. For each entry $q^t{}_i$, the decoder predicts its bounding box coordinates $bb^t{}_i$, containing the object's center coordinates, width, and height. The center coordinates denote the position of the object within the frame, while the width and height provide information about the object's size. This approach enables the simultaneous solution of object detection and data association problems. During inference, entries with $s^t{}_i \geq \varepsilon$ are retained by applying a threshold to each entry in $[Q^t{}_{can}, Q^t{}_{tracklet}]$. The resulting entries will either inherit a track identity or initiate a new track based on whether they come from $'Q^t{}_{can}$ or $'Q^t{}_{tracklet}$. The final tracking results are obtained by merging track identities, either inherited or new, with the corresponding bounding box predictions, without the need for further post-processing [3, 16]. Supervision signals for $o^t{}_i$, $u^t{}_i$, and $bb^t{}_i$ are generated for each frame by first assigning objectiveness scores and bounding boxes to entries in $'Q^t{}_{tracklet}$, depending on the presence of the tracked object in the frame. For each entry in $'Q^t{}_{can}$, ground truth bounding boxes, whether these are newly identified or already being tracked, are assigned through bipartite matching [19]. Bipartite matching assigns ground truth bounding boxes to each entry in $Q^t{}_{can}$ with the objective to find a maximum matching in a bipartite graph (a graph whose vertices can be divided into two disjoint sets) that covers all vertices of one set. Subsequently, each proposal gets assigned a uniqueness score based on prior object encounters, as illustrated in Fig.8.

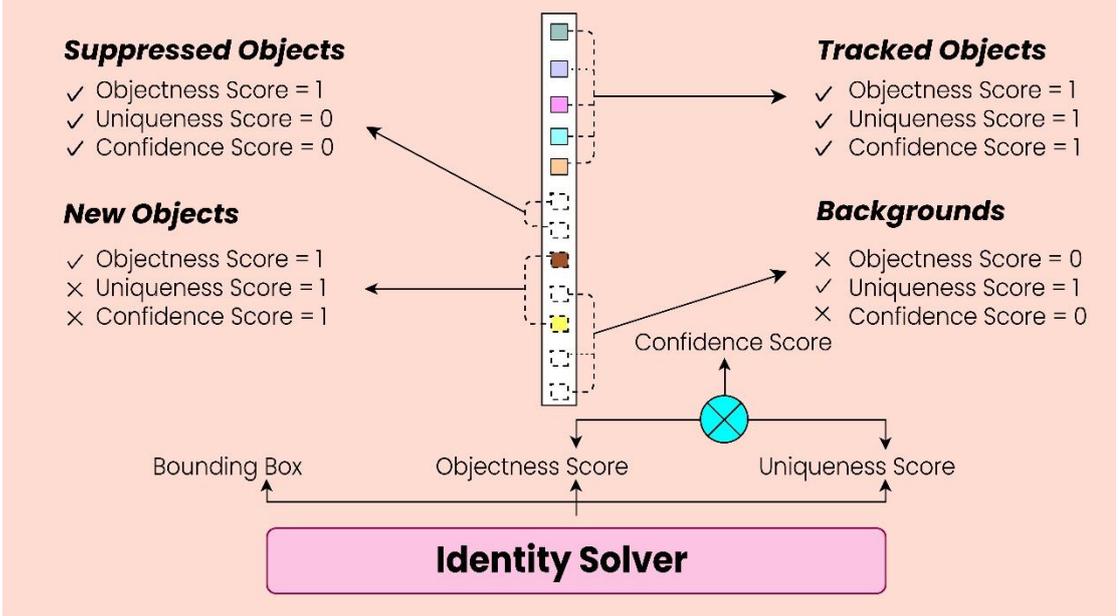

Figure 5.6 Association Assignment Workflow

## 5.7. LOSS FUNCTIONS

In training STMMOT, we compute the tracking loss based on three components objectness score $o^t{}_n$, uniqueness score $u^t{}_n$, and bounding box transformation $bb^t{}_n$ using the assignment procedure described earlier. The tracking loss $L_{track}$ is defined as:

$$L_{track} = \lambda_{cls}(L_{obj} + L_{uni}) + \lambda_{L1}(L_{bbox} + L_{iou}) \quad (7)$$

where $\lambda$ represents the weight scaling factor that ensures appropriate balance among the different loss components, $L_{obj}$ and $L_{uni}$ are focal losses for objectness and uniqueness scores, $L_{bbox}$ is the L1 loss for bounding box regression, and $L_{iou}$ is the generalized IoU loss. The assignment of object instances to these components is carried out as it typically is in standard object detection tasks, and the corresponding loss is computed as:

$$L_{det} = \lambda_{cls}L_{obj} + \lambda_{L1}L_{bbox} + \lambda_{iou}L_{iou} \quad (8)$$

Post-training, the auxiliary decoder is detached. Inspired by the MOTR

approach [26], the tracking loss for a series of frames is computed as the cumulative loss across individual track queries, with this sum being normalized by N instances. For a sequence that includes T frames, the total sequence loss, denoted as $L_{seq}$, is a fusion of the tracking and the auxiliary detection loss:

$$L_{seq} = \lambda_{track} L_{seq-track} + \lambda_{det} L_{seq-det} \tag{9}$$

where $\lambda_{track} \in R$ and $\lambda_{det} \in R$ are the weights factors. $\lambda_{track}$ specifically determines the relative importance of tracking loss in our overall cost function. The tracking loss considers the model's ability to consistently follow objects across successive frames, ensuring continuity in object tracking. If the model loses track of an object from one frame to the next, this would increase the tracking loss. Thus, a higher $\lambda_{track}$ gives more weightage to this aspect of the loss function, prioritizing consistency in object tracking across frames. Similarly, $\lambda_{det}$ is the weight assigned to the auxiliary detection loss. This loss component accounts for the model's ability to correctly identify and localize new objects in each frame. This involves accurately determining the bounding box dimensions and the class of each detected object. A larger $\lambda_{det}$ value implies more emphasis is being placed on the detection aspect of the model's performance. Both $\lambda_{track}$ and $\lambda_{det}$ are tuned during the training process, optimizing the balance between the tracking and detection capabilities of our model. The selection of these weights is largely dependent on the specific application and the data at hand. Finally, $N_t$ is a variable that signifies the visible objects present in a frame at a given time t. This variable is crucial because it influences the normalization of the tracking loss over a sequence of frames. Essentially, the cumulative tracking loss is divided by $N_t$, to calculate the average loss per visible object in the frame. This ensures that our model's performance evaluation is not unfairly penalized when more objects are present in a frame.

# 6.   EXPERIMENTAL SETUP

In order to validate the efficiency of the technique, we perform extensive testing on two renowned pedestrian tracking datasets, namely, MOT17 [26] and MOT20 [7]. In our ablation study, we follow the established procedure [58] by dividing the MOT17 training set into two sections, one for training and the other for validation. We utilize the widely-accepted MOT metrics set [2] for quantitative assessment, with multiple object tracking accuracy (MOTA) serving as the primary metric for gauging overall performance.

## 6.1.   DATASET

### 6.1.1.   Crowd Human

DeepSportradar-ReID [59] is a unique dataset derived from short tracking sequences of basketball games. Each sequence within the dataset comprises 20 frames, making it suitable for multi-frame analysis tasks. The dataset consists of a total of 38 sequences featuring 486 unique individuals, each meticulously annotated with attributes like person identity, bounding box coordinates, identity numbers (IDs), and camera information. One of the notable aspects of the DeepSportradar-ReID dataset is the variability in image resolution across sequences. This variability is reflective of the unpredictable and dynamic nature of real-world environments and presents researchers with realistic challenges tied to resolution variations in the field of computer vision. Furthermore, this dataset is collected using multiple cameras placed in various positions around the basketball games, providing an authentic and diverse range of perspectives. The diversity in camera angles and perspectives contributes to the robustness of the dataset, making it ideal for developing and testing algorithms intended to operate in complex, real-world scenarios. Overall, the DeepSportradar-ReID dataset provides a valuable resource for advancing research in person re-identification and multi-object tracking in sports analytics.

### 6.1.2. MOT17

The MOT17 [54] dataset concentrates on tracking multiple individuals in crowded scenarios. It contains a total of 14 video sequences, with seven designated for testing. The MOT17 is widely utilized to benchmark MOT methodologies [3, 47, 51]. Following previous studies [37, 43, 45], we divide the MOT17 into two subsets during validation. We employ the first subset for training and the second for validation purposes.

### 6.1.3. MOT20

The MOT20 [55] dataset comprises 8 demanding video sequences set in uncontrolled environments and crowded scenes. It is divided into two parts: 4 sequences for the training set and 4 for the testing set. We train on the MOT20 training partition using identical hyper-parameters as those employed for the MOT17 dataset.

### 6.2. EVALUATION METRICS

The Multi-Object Tracking (MOT) benchmark [54, 55] employs Multi-Object Tracking Accuracy (MOTA) as its primary metric. MOTA is defined as:

$$MOTA = 1 - (\Sigma_n(FP_n + FN_n + IDSW_n)) / \Sigma_n(GT_n) \ (10)$$

where $GT_n$ denotes the number of ground truth objects in the nth frame, while $FP_n$, $FN_n$, and $IDSW_n$ represent the errors associated with false positives (FP), false negatives (FN), and ID switches (IDSW), respectively. Additionally, following the MOT benchmark recommendations, we report HOTA, a novel tracking metric [28]. HOTA is the geometric mean of Detection Accuracy (DetA) and Association Accuracy (AssA). AssA follows the formula |TP| / (|TP| + |FN| + |FP|), with their specific true/false criteria. In our experience

DF1 (Identity F1 Score) Measures the accuracy of object identity preservation. IDF1 = 2 * (Precision * Recall) / (Precision + Recall)

FP is the number of false detections that were not associated with a ground truth target. FN is the number of ground truth targets that were not detected by the tracker. TP refers to the number of correctly identified targets by the re-identification system. IDSW is the number of times that the tracker switches the identity of a target with another target.

### 6.3.  EXPERIMENT SETTINGS

Our proposed method is developed using PyTorch, with training and validation executed on a machine with a 16-core CPU @ 3.60 GHz, supported by an Nvidia GeForce RTX 4090 GPU having 128 GB RAM. Data augmentation techniques, such as random horizontal flips, random crops, and scale augmentation, are employed, resizing input images so their shorter side is 900 pixels and the longer side is no more than 1280 pixels. We utilize EfficientNet [57] as the network backbone and DETR [27] pre-trained on MSCOCO [56] for candidate proposal creation. All Transformer units have their number of layers reduced to 4. Our memory buffer can hold up to 350 tracks for the MOT17 and 700 tracks for MOT20.  The maximum temporal length is 30 for MOT17 and 35 for MOT20, mainly limited by GPU memory constraints. This length refers to the maximum sequence of frames that STMMOT processes at once. The transformer has been assigned an initial learning rate of 0.03 (3e-2), while the backbone's learning rate has been set at a much lower value of 0.00002 (2e-5). The choice of these specific learning rates is determined by the different roles these modules play within the network architecture. To prevent overfitting and ensure a more stable training process, a weight decay parameter is incorporated, which is set to 0.01 (1e-2).

This regularization technique helps in avoiding large weights, thus leading to a simpler and more generalizable model. As for weight initialization, the transformer weights are assigned initial values using Xavier initialization, and the backbone model is pre-trained with frozen batch-norm layers [58]. In line with previous work [19, 27], we choose the coefficients of the Hungarian loss with $\lambda_{cls}, \lambda_{L1}$ and $\lambda_{iou}$ as 3, 6, and 3, respectively. We set $\lambda_{det} = \lambda_{track} = 2$ in Eq. 9. The model is trained for 200 epochs, with the learning rate decreasing by a factor of 10 at the 100th epoch. We have employed a sequence-centric training approach. We utilize a sequence-oriented training scheme where the initial sequence length starts at 4 and grows by an increment of 4 for every 20 epochs. The frames within each sequence are chosen using a random interval that can vary between 1 and 10. A notable aspect of our approach, STMMOT, is that, unlike other leading-edge methods, it is trained on both the CrowdHuman training and validation sets and the MOT17 training set. In the case of the MOT20 benchmark, no supplementary data sources were leveraged.

# 7.    RESULTS AND DISCUSSION

we perform extensive testing on two renowned pedestrian tracking datasets, namely, MOT17 [26] and MOT20 [7]. In our ablation study, we follow the established procedure [58] by dividing the MOT17 training set into two sections, one for training and the other for validation. We utilize the widely-accepted MOT metrics set [2] for quantitative assessment, with multiple object tracking accuracy (MOTA) serving as the primary metric for gauging overall performance. Figure 9 and Figure 10 visualize some tracking results of STMMOT on MOT17 and MOT20. The correct assignment of IDs, despite the challenges posed by occlusion or object reappearance. Moreover, STMOT shows robust detection of smaller objects, correctly identifying their respective IDs, which exhibits the strength of CPN. The experimental findings reveal that the proposed STMMOT is capable of unbroken tracking of the target, maintaining a steady identity label when the target re-emerges post-occlusion. This outcome underscores the robust tracking capabilities of the proposed method in overcoming in scenarios characterized by the high density of tracking targets and temporal occlusion.

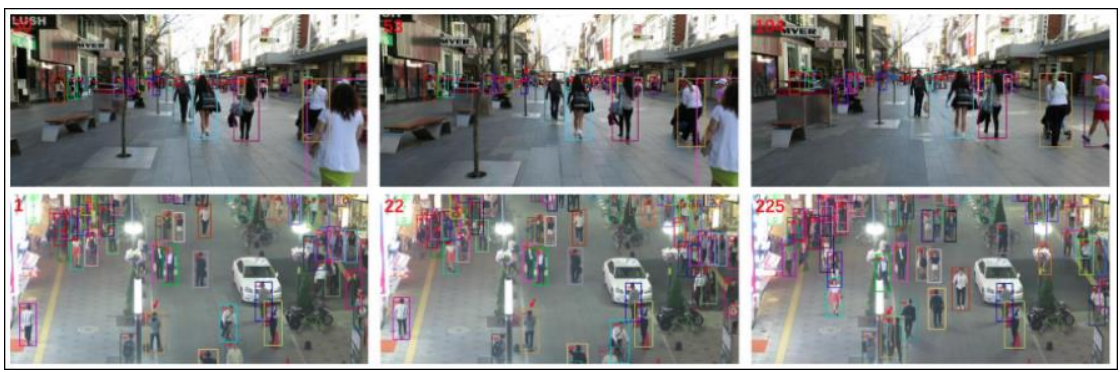

Figure 7.1 Sample tracking results of STMMOT on the MOT17.

## 7.1.  ABLATION STUDY

We carry out tests with different Transformer architectures, configurations of memory, PFTL units, and model structures. Unless specifically stated, our models are streamlined by decreasing the layer count of all Transformer units from 4 to 2. The training is carried out on the CrowdHuman and MOT17 datasets, and we validate the models using the MOT17 validation dataset while ensuring that any overlapping videos from the training set are not included in the validation set.

### 7.1.1. Comparison of Transformer Architecture

We conduct an evaluation to understand the impact of various Transformer architectures. Five such structures are examined. The first, referred to as Transformer,

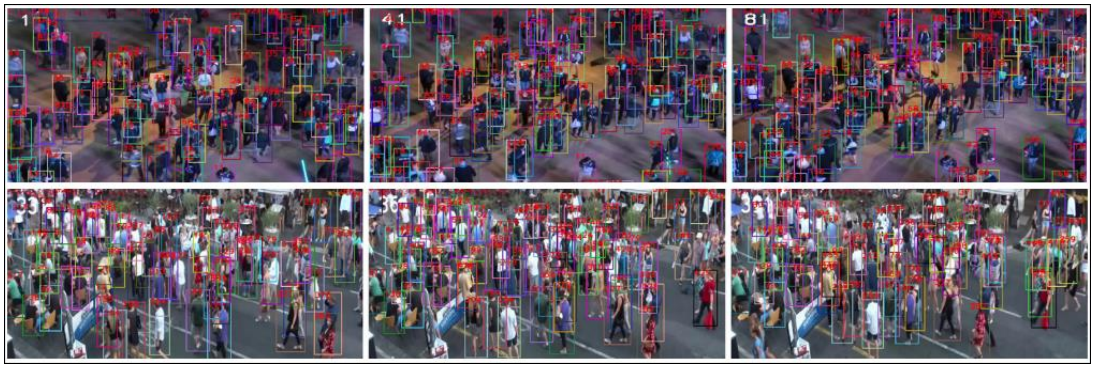

Figure 7.2 The MOT example tracking results of the proposed STMOT on MOT20 sequences which shows that STMOT can track multiple objects in dense and occluded scenes while achieving excellent performance.

Employs the design from DEtection TRansformer (DETR) [60] where the transformer is built atop the feature maps from the ResNet 5th stage [61]. The second structure, DETR-DC5 [60], enhances the resolution of these feature maps through dilation convolution applied to the ResNet 5th stage and removing a stride from the initial convolution of this stage. The third structure, DETR-FPN, implements FPN [22] on the input feature maps. The encoder of the Transformer is entirely removed from the pipeline due to memory constraints, which in turn allows the learning rate of the backbone to be increased to match that of the transformers. Lastly, the Deformable Transformer [27], a recently suggested architecture to address the limited resolution issue in transformers, is explored. It incorporates multiple-scale features into the entire encoder-decoder pipeline, which has proven effective on general object detection datasets within acceptable memory usage parameters.

Table 1 presents the quantitative results. The overall performance of baseline DETR is relatively low because of low feature resolution at IDF1 score of 57.8, MOTA of 59.4, and HOTA of 51.2. The DETR-DC5 which improves the feature resolution, shows an improvement across all metrics when compared to the baseline DETR model. The IDF1 score increases by 6.7 to 64.5, MOTA improves by 4.3 to 63.7, and HOTA and AssA show increases of 4.4 and 7.5 to 55.6 and 49.6, respectively. In terms of error metrics, IDSW decreases by 1.9 to 29.6, FP decreases by 2.3 to 6.2, and FN decreases by 0.5 to 1.6. However, it also has a drawback in terms of high memory usage due to dilation convolution. The DETR-FPN model exhibits a notable increase in performance compared to both the DETR and DETR-DC5 models. The IDF1 score decreases slightly by 1.7 to 62.8 compared to DETR-DC5, but MOTA increases significantly by 4.4 to 68.1. DETR-FPN's performance is comparable to DETR-DC5, indicating that a resolution higher than DC5 does not necessarily result in significant performance improvement, possibly due to the absence of encoder blocks. The Deformable Transformer, with its fusion of multiple-scale features into the full encoder-decoder pipeline, stands out with the most significant performance increase. The IDF1 score increases by 11.0 to 73.8 compared to the DETR-FPN model. MOTA shows an increase of 9.2 to 77.3, while HOTA and AssA improve to 62.2 and 61.2, respectively. Notably, error metrics decrease substantially, with IDSW decreasing by 4.1 to 24.2, FP decreasing by 1.8 to 4.1, and FN decreasing by 0.9 to 0.3. These comparisons demonstrate the effectiveness of the Deformable Transformer architecture in improving STMMOT performance across all metrics. Consequently, we select Deformable Transformer as our default architecture choice for STMMOT.

Table 7.1 Ablation Study on Different Transformer Based Different Candidate Generator

| CPN Architecture | IDF1 ↑ | MOTA ↑ | HOTA ↑ | AssA↑ | IDSW (%) ↓ | FP (%) ↓ | FN (%) ↓ |
|---|---|---|---|---|---|---|---|
| DETR [60] | 57.8 | 59.4 | 51.2 | 42.1 | 31.5 | 8.5 | 2.1 |
| DETR-DC5 [60] | 64.5 | 63.7 | 55.6 | 49.6 | 29.6 | 6.2 | 1.6 |
| DETR-FPN | 62.8 | 68.1 | 59.1 | 53.7 | 28.3 | 5.9 | 1.2 |
| Deformable Transformer [27] | 73.8 | 77.3 | 62.2 | 61.2 | 24.5 | 4.1 | 0.3 |

### 7.1.2. Comparison of Backbone Networks

We undertake this ablation to evaluate the impact of various backbone networks used as feature extractors for the Deformable Transformer in candidate generation. We replace EfficientNet-B4 [63], the default feature extractor with various backbones including Res2Net[65], DenseNet-201[66], Inception-v3[67], and ResNet-50[64]. For each stream in these backbones, we discard the classifiers, comprised of two linear layers. As Table 2 shows, Res2Net has a total of 25.5 million parameters and its performance shows an IDF1 score of 65.3, MOTA of 66.1, HOTA of 56.1, and AssA of 57.8. Next, we employed DenseNet-201, which consists of 20 million parameters. Its performance exhibits an IDF1 score of 67.7, MOTA of 67.3, HOTA of 58.5, and AssA of 57.3. It demonstrates a slight increase in ID switches and false positives to 26.7% and 5.4% respectively, and a marginal increase in false negatives to 0.7%. The ResNet-50 backbone, having 25.6 million parameters, shows a significant improvement with an IDF1 score of 73.6, MOTA of 75.4, HOTA of 61.1, and AssA of 60.4. It further decreases ID switches to 24.3% and false positives to 4.9%, however, false negatives slightly increased to 0.8%. Finally, using EfficientNet-B4, with 19 million parameters, the IDF1 score is 73.8, MOTA is 77.3, HOTA is 62.2, and AssA is 61.2. It slightly increases ID switches to 24.5%, but decreases false positives to 4.1% and false negatives to 0.3%." We select EfficientNet [63] as the default feature extractor for STMMOT because of its high performance on the majority of metrics and less number of parameters.

Table 7.2 Ablation Study on Different Backbone Networks as Feature Extractor for Deformable Transformer to Generate Candidates

| Backbone | Params (M) | IDF1 ↑ | MOTA ↑ | HOTA ↑ | AssA↑ | IDSW (%) ↓ | FP (%) ↓ | FN (%) ↓ |
|---|---|---|---|---|---|---|---|---|
| Res2Net [65] | 25.5 | 65.3 | 66.1 | 56.1 | 57.8.1 | 26.7 | 5.1 | 0.6 |
| DenseNet-201 [66] | 20 | 67.7 | 67.3 | 58.5 | 57.3 | 26.7 | 5.4 | 0.7 |
| Inception-v3 [67] | 23.9 | 71.4 | 71.8 | 58.2 | 58.0 | 26.4 | 5.0 | 0.7 |
| ResNet-50 [64] | 25.6 | 73.6 | 75.4 | 61.1 | 60.4 | 24.3 | 4.9 | 0.8 |
| EfficientNet-B4 [63] | 19 | 73.8 | 77.3 | 62.2 | 61.2 | 24.5 | 4.1 | 0.3 |

### 7.1.3. Comparison of Short-term Memory Length

STMMOT utilizes the short-term memory buffer to handle the non-linear movement problem and mitigate the tracking problem due to short-term appearance-reappearance. This ablation study analyzes the impact of varying the short-term memory length in the Deformable Transformer, where EfficientNet-B4 is used as the backbone network and long-term memory TLTM is kept at 25. The experiment starts with a short-term memory length of 2, with a long-term memory length of 30, the metrics show an IDF1 score of 70.2, MOTA of 74.7, HOTA of 60.2, and AssA of 59.4. When the short-term memory length is increased to 3, there is a notable improvement in all metrics: IDF1 increases to 72.4, MOTA to 75.1, HOTA to 61.0, and AssA to 61.0. It also results in a decrease in ID switches to 25.9% and false positives to 4.9%, while significantly reducing false negatives to 0.4%. The optimal results are observed with a short-term memory length of 5: IDF1 reaches 73.8, MOTA hits 77.3, HOTA is 62.2, and AssA is 61.2. At this point, ID switches and false positives fall to their lowest at 24.2% and 4.1% respectively, while false negatives remain at 0.3%. However, when the short-term memory length is increased to 6 or 7, the metrics either stagnate or slightly degrade. For instance, the IDF1 score dips to 73.2 and then to 72.8, while MOTA maintains at 77.3 and then slightly drops to 77.0. HOTA slightly decreases to 62.0 and then to 61.6, and AssA drops to 60.7 and further to 60.1. Observation in Table 3 tells that there is a significant improvement in short-term memory length from 2 to 5, however, no significant performance improvement on increasing the length further, rather than starts decreasing. It can be established that higher memory length improves performance, but hardware limitations can be the hurdles in selecting the optimal length. Hence, length 5 is a trade-off between performance and computational efficiency, and it is selected as the default short-term memory length for the STMMOT.

Table 7.3 Ablation Study on Varying Short-Term Memory Length in Deformable Transformer with EfficientNet-B4 as Backbone Network

| $T_{STM}$ | $T_{LTM}$ | IDF1 ↑ | MOTA ↑ | HOTA ↑ | AssA↑ | IDSW (%) ↓ | FP (%) ↓ | FN (%) ↓ |
|---|---|---|---|---|---|---|---|---|
| 2 | 25 | 70.2 | 74.7 | 60.2 | 59.4 | 26.7 | 5.1 | 0.7 |
| 3 | | 72.4 | 75.1 | 61.0 | 61.0 | 25.9 | 4.9 | 0.4 |
| 4 | | 73.1 | 75.4 | 61.7 | **61.8** | 25.3 | 4.5 | 0.4 |
| 5 | | **73.8** | **77.3** | **62.2** | 61.2 | **24.2** | **4 .1** | **0.3** |
| 6 | | 73.2 | **77.3** | 62.0 | 60.7 | 24.8 | 4.7 | **0.3** |
| 7 | | 72.8 | 77.0 | 61.6 | 60.1 | 25.8 | 5.0 | 0.5 |

### 7.1.4. Comparison of Long-term Memory Length

Employing a long-term memory buffer, STMMOT effectively addresses the common occurrence of occlusion in real-world contexts. Table 4 compares the impact of different long-term memory lengths while keeping the length for short-term memory at 5. Increasing the memory from 5 to 25, there is a significant improvement in the performance which is quite obvious from the reduction in ID switching. The best performance occurs with a long-term memory length of 25: the IDF1 reaches 73.8, MOTA and HOTA peak at 77.3 and 62.2 respectively, and AssA hits 61.1. Furthermore, ID switches and false positives decrease to their lowest at 24.2% and 4.1%, respectively, while false negatives remain at 0.3%. However, when the long-term memory length is extended to 30, most metrics show a slight increase in performance. Based on these trends, it can be inferred that a long-term memory length of 25 provides the most optimal performance and provides the best trade-off for accuracy-efficiency given our hardware limitation.

Table 4: Ablation study on varying long-term memory length in Deformable Transformer with EfficientNet-B4 as a backbone network

| $T_{STM}$ | $T_{LTM}$ | IDF1 ↑ | MOTA ↑ | HOTA ↑ | AssA ↑ | IDSW (%) ↓ | FP (%) ↓ | FN (%) ↓ |
|---|---|---|---|---|---|---|---|---|
| 5 | 5 | 71.1 | 74.4 | 60.0 | 59.4 | 27.1 | 5.7 | 0.6 |
| | 10 | 71.3 | 76.4 | 60.2 | 60.4 | 26.5 | 4.8 | 0.6 |
| | 15 | 71.8 | 77.1 | 61.4 | 60.1 | 26.3 | 4.9 | 0.4 |
| | 20 | 72.6 | 77.1 | 62.4 | **61.3** | 25.4 | 4.9 | **0.3** |
| | 25 | **73.8** | **77.3** | **62.2** | 61.1 | **24.2** | **4.1** | **0.3** |
| | 30 | **73.9** | 77.4 | 61.7 | 61.1 | **23.9** | 4.0 | 0.4 |

### 7.1.5. Comparison of Memory Aggregation Designs

We experiment with the different structure of the memory aggregation module by initiating comparisons with heuristic pooling and two alternative attention-based aggregation structures into memory encoding. Given that the length of the tracklet can extend up to 25, we choose not to concatenate the embeddings, instead opting to evaluate pooling methodologies. The aggregation could be executed through either the calculation of the arithmetic mean or the maximum norm, encompassing the most recent T frames. As depicted in Table 5, it becomes clear that these basic pooling methods fail to capture the informative track features, leading to a significant reduction in IDF1 and MOTA performance as compared to the default aggregation method of STMMOT. The first attention-based aggregation structure uses only a cross-attention module, excluding the distinction between long and short-term memory. This strategy employs the most recent observation to query an object's past T embeddings. For the Single aggregation method, with T=5, we observed significant improvements with an IDF1 score of 72.6, MOTA of 68.7, and HOTA of 58.9. With a memory length of 25 (T=25), the performance was slightly decreased compared to T=5, but it remained competitive. Taking inspiration from LSTR [70], the second method utilizes aggregated short-term embeddings to extract valuable information from long-term memory. The mixed aggregation method 'Long-after-short' showed overall worse performance compared to the Single method, but our proposed design outperformed all other methods. We propose that in the task of action detection, LSTR's main concentration, each frame's outcome is independent, and any shortcomings in short-term features have a minor effect on future predictions. However, when it comes to MOT, errors in associations can have a cumulative effect, which underscores the efficiency of using long-term features to make up for any weakness in short-term features.

Table 7.4 Ablation Study on Different Memory Aggregation Designs in Deformable Transformer with EfficientNet-B4 as Backbone Network

| Design | Memory Length | IDF1 ↑ | MOTA ↑ | HOTA ↑ | AssA↑ | IDSW (%) ↓ | FP (%) ↓ | FN (%) ↓ |
|---|---|---|---|---|---|---|---|---|
| Pooling | Avg T=5 | 58.7 | 44.6 | 41.9 | 47.9 | 29.7 | 4.5 | 0.9 |
| | Max T=5 | 54.1 | 37.0 | 34.1 | 47.5 | 31.5 | 7.2 | 1.4 |
| | Avg T= 25 | 38.4 | 28.6 | 30.0 | 34.5 | 27.0 | 8.8 | 2.1 |
| | Max T= 25 | 47.0 | 33.4 | 31.7 | 42.8 | 34.4 | 7.3 | 2.3 |
| Single | $T = 5$ | 72.6 | 68.7 | 58.9 | 59.5 | 28.4 | **3.9** | **0.4** |
| | $T = 25$ | **72.8** | 66.25 | **58.7** | **60.2** | **26.1** | 5.2 | 0.8 |
| Long-after-short-Memory | - | 71.4 | **67.4** | 54.0 | 57.4 | 32.4 | 6.4 | 1.7 |
| Ours | - | **73.8** | **77.3** | **62.2** | **61.2** | **24.2** | **4.1** | **0.3** |

### 7.1.6. Design Comparison of the Number of PFTL

We perform the ablation study to show the effectiveness of PFTL and the number of PFTL suitable that balance the performance and computational efficiency. We have compared the performance of base STMMOT without PFTL and with different numbers of PFTL. Initially, the STMMOT model without PFTL achieved an IDF1 score of 67.2%, MOTA of 70.7%, and HOTA of 52.3%. The model also demonstrated an association accuracy (AssA) of 52.7%, with identity switches (IDSW) at 29.5%, false positives (FP) at 6.7%, and false negatives (FN) at 0.6%. When PFTL was introduced at Level 1, the model exhibited improvements across the board. IDF1 increased to 71.1% (a 5.8% improvement), MOTA to 72.4% (a 2.4% boost), and HOTA to 55.9% (a 6.9% uplift). AssA rose to 56.6% (a 7.4% increase), while the IDSW, FP, and FN percentages reduced to 28.1% (a decrease of 4.7%), 5.3% (down by 20.8%), and 0.6% (no change), respectively. On reaching PFTL 4, the STMMOT model's performance further improved. IDF1 rose to 73.8% (an increase of 3.8%), MOTA to 77.3% (up by 6.7%), and HOTA to 62.2% (up by 11.2%). Finally, at PFTL 7, the STMMOT model achieved its peak performance. IDF1 was marginally better at 74.2% (up by 0.5%), MOTA improved to 79.3% (up by 2.6%), and HOTA went up to 64.4% (an increment of 3.5%). It can be observed that the sequential addition of PFTLs in STMMOT from 1 to 7 progressively improves the model's performance. The model seems to reach its peak performance at PFTL 7, albeit with only marginal improvements from 4 PFTLs. To better trade-off between performance and computational efficiency given the hardware limitations, we use the 4 PFTLs at each level of the SVP which is used in the rest of the experiment.

Table 7.5 Ablation Experiment on STMMOT without PFTL and Different PFTLs

| Design | PFTL | IDF1 ↑ | MOTA ↑ | HOTA ↑ | AssA↑ | IDSW (%) ↓ | FP (%) ↓ | FN (%) ↓ |
|---|---|---|---|---|---|---|---|---|
| STMMOT without PFTL | - | 67.2 | 70.7 | 52.3 | 52.7 | 29.5 | 6.7 | 0.6 |
| STMMOT With PFTL | 1 | 71.1 | 72.4 | 55.9 | 56.6 | 28.1 | 5.3 | 0.6 |
| | 2 | 71.8 | 73.7 | 59.5 | 59.8 | 27.8 | 5.2 | 0.5 |
| | 3 | 72.4 | 75.9 | 51.7 | 60.1 | 25.7 | 4.7 | 0.4 |
| | **4** | 73.8 | 77.3 | 62.2 | 61.2 | 24.2 | 4.1 | 0.3 |
| | 5 | 73.9 | 78.1 | 63.1 | 61.7 | 24.1 | 4.0 | 0.3 |
| | 6 | 74.1 | 78.8 | 63.9 | 62.1 | 24.1 | 3.9 | 0.3 |
| | **7** | 74.2 | 79.3 | 64.4 | 62.4 | 23.9 | 3.9 | 0.3 |

## 7.2. PERFORMANCE COMPARISON

For a fair comparison, we emphasize contrasting our approach, STMMOT, primarily with methodologies that implement an in-network identity association solver (IIAS). The IIAS provides a mechanism for predicting object identities directly within the network, effectively eliminating the necessity for any post-processing stage. This kind of arrangement offers certain advantages, such as increased efficiency and streamlined computation, since the entire operation can be embedded into the network and performed in an end-to-end manner. On the other hand, some techniques utilize a post-network identity association solver (PIAS). This approach applies various rule-based linking procedures to the detected results. Common methods include the application of Hungarian matching algorithms paired with Kalman Filters and re-ID features. The Kalman filter, on the other hand, is a recursive state estimation algorithm, typically utilized to predict the future state of a tracked object based on its past states. Despite their potential effectiveness, such heuristic linking strategies can impose limits on scalability and general applicability. This is primarily due to the reliance on post-processing and manually curated rules, which can become computationally intensive as the number of objects increases. Furthermore, these methods may not adapt well to variations in tracking scenarios, making them less flexible and scalable compared to methods employing an IIAS.

### 7.2.1. MOT17 Benchmark

Table 7 demonstrates that STMMOT attains better performance among IIAS MOTs, while comparative performance is compared to PIAS methods. Performance comparison of STMMOT against two powerful models, namely StrongSort and TransMOT, offers valuable insights. StrongSort demonstrates remarkable performance across several benchmarks. However, STMMOT still manages to outperform

StrongSort on multiple measures. For instance, the IDF1 score sees an increase from 78.5% with StrongSort to 79.8% with STMMOT, suggesting enhanced identity preservation in the latter. Similarly, the MOTA score rises by 2.0 percentage points, from 78.3% in StrongSort to 79.3% in STMMOT, demonstrating the superior tracking accuracy of our proposed model. Particularly noteworthy is the significant increment in the HOTA score, which surges by 9.7 percentage points from 63.5% in StrongSort to a robust 73.2% in STMMOT, reflecting a substantial enhancement in both detection and association accuracy. Although AssA shows a slight decrement and the number of identity switches (IDSW) increases, STMMOT maintains a competitive edge in these areas. When compared with TransMOT, STMMOT maintains its performance edge. Specifically, the IDF1 score sees an uplift of 3.5 percentage points from 76.3% with TransMOT to 79.8% in STMMOT, once again underlining the model's robustness in preserving object identities. Similarly, the MOTA score witnesses a modest increment of 2.9 percentage points, rising from 76.4% with TransMOT to 79.3% with STMMOT, reinforcing STMMOT tracking accuracy. The IDSW value shows a significant decrement from 1623 with TransMOT to 1529 with STMMOT, suggesting that STMMOT achieves more accurate data association by reducing identity switches.

Table 7.6 Performance Comparison of MOT Models on MOT17 Test Set

| Method | Year | Transformer | IIAS | Memory Network | Scale Variant | IDF1 ↑ | MOTA ↑ | HOTA ↑ | AssA↑ | IDSW ↓ |
|--------|------|-------------|------|----------------|---------------|--------|--------|--------|-------|--------|
| FairMOT [16] | 2021 | - | - | - | - | 72.3 | 73.7 | 59.3 | 58.0 | 3303 |
| CenterTrack [14] | 2020 | - | - | - | - | 64.7 | 67.8 | | | 2583 |
| CenterTrack-MOTSynth No [68] | 2021 | - | - | - | - | 52.0 | 59.7 | - | - | 6035 |
| StrongSort [3] | 2023 | - | - | - | - | **78.5** | **78.3** | **63.5** | **63.7** | 1446 |
| TDT [69] | 2022 | - | - | - | ✓ | 60.9 | 63.8 | - | - | 4401 |
| Transtrack [29] | 2020 | ✓ | - | - | ✓ | 63.5 | 75.2 | 54.1 | 47.9 | 4614 |
| TransCenter [28] | 2021 | ✓ | - | - | - | 62.2 | 73.2 | 54.5 | 49.7 | 3663 |
| MOTR [26] | 2021 | ✓ | ✓ | - | - | 67.0 | 67.4 | - | - | 1992 |
| MeMOT [45] | 2022 | ✓ | ✓ | ✓ | - | 69.0 | 72.5 | 56.9 | **55.2** | 2724 |
| TrackFormer [25 ] | 2022 | ✓ | ✓ | - | - | 63.9 | 65.0 | - | - | 3258 |
| TransMOT [37 ] | 2023 | ✓ | ✓ | - | - | **76.3** | **76.4** | - | - | **1,623** |
| locality-enhanced [73] | 2023 | ✓ | - | - | ✓ | | 72.1 | - | - | 2087 |
| Swin-JDE [ 31] | 2023 | ✓ | ✓ | - | - | 70.7 | 72.3 | **57.8** | - | 2679 |
| STMMOT (ours) | | ✓ | ✓ | ✓ | ✓ | **79.8** | **79.3** | **73.2** | **61.2** | **1529** |

### 7.2.2. MOT20 Benchmark

We evaluated our proposed model, STMMOT, against a variety of state-of-the-art multiple object tracking (MOT) models on the MOT20 test set. As Table 8 shows, FairMOT model, with an IDF1 score of 67.3 and MOTA of 61.8, had a considerable number of identity switches (IDSW) at 5243. On the other hand, StrongSort showed improved performance with an IDF1 score of 73.8 and MOTA of 77.3, significantly reducing the IDSW count to 1729. ByteTrack and TransMOT achieved higher performance with IDF1 scores of 75.1 and 75.2 and MOTA scores of 76.5 and 77.4, respectively. However, the number of IDSW for ByteTrack was 1120, and for TransMOT, it was 1601. STMMOT outperforms all other models, achieving the highest IDF1 score of 78.4, MOTA score of 74.1, and keeping the IDSW count at a low of 1264. Additionally, it recorded high HOTA and AssA scores, 69.0 and 61.5, respectively. The superior performance of STMMOT can be attributed to the model's implementation of transformer architectures and memory networks, a combination seen in only a few of the compared models. The inclusion of scale variant designs in STMMOT might have also contributed to its dominant performance.

Table 8: Performance comparison of MOT models on MOT20 test set

| Method | Year | Transformer | IIAS | Memory Network | Scale Variant | IDF1 ↑ | MOTA ↑ | HOTA ↑ | AssA↑ | IDSW ↓ |
|--------|------|-------------|------|----------------|---------------|--------|--------|--------|-------|--------|
| FairMOT [16] | 2021 | - | - | - | - | 67.3 | 61.8 | 54.6 | 54.7 | 5243 |
| StrongSort [3] | 2023 | - | - | - | - | **73.8** | 77.3 | **62.2** | **61.2** | 1729 |
| CenterTrack-MOTSynth No [68] | 2021 | - | - | - | - | 39.7 | 43.7 | - | - | 3467 |
| TDT [69] | 2020 | - | - | - | ✓ | 46.0 | 47.9 | - | - | 5342 |
| ByteTrack [5] | 2022 | - | - | - | - | 75.1 | **76.5** | 61.2 | 60.0 | **1,120** |
| Transtrack [29] | 2020 | ✓ | - | - | ✓ | 59.4 | 65.0 | 48.9 | 45.2 | 3608 |
| TransCenter [28] | 2021 | ✓ | - | - | - | 49.6 | 58.5 | 43.5 | 37.0 | 4695 |
| MeMOT [45] | 2022 | ✓ | ✓ | ✓ | - | 66.1 | 63.7 | 54.1 | **55.0** | 1938 |
| TrackFormer [25 ] | 2022 | ✓ | ✓ | - | - | 63.6 | | | | |
| TransMOT [37] | 2023 | ✓ | ✓ | - | - | **75.2** | **77.4** | - | - | **1,601** |
| Swin-JDE [31 ] | 2023 | ✓ | ✓ | - | - | 69.5 | 70.4 | **55.7** | - | 2026 |
| STMMOT (ours) | | ✓ | ✓ | ✓ | ✓ | **78.4** | **74.1** | **69.0** | **61.5`** | **1264** |

# 8.   CONCLUSION AND FUTURE WORK

## 8.1.   CONCLUSION

Our STMMOT proposition addresses online MOT through the integration of object detection and identity association. By integrating the strengths of transformer architectures and memory networks, our model successfully manages the complexity and uncertainties that come with tracking multiple objects in real-world environments. STMMOT maintains a broad spatio-temporal memory and translates previous observations using a memory aggregator that is attention-based. Through dynamically updating query embeddings that represent objects, STMMOT is capable of forecasting the states of these objects using an attention mechanism, eliminating the need for any post-processing. The results from our extensive experimentation and ablation studies validate the effectiveness and robustness of STMMOT under different settings and conditions. Key to the success of our model is the implementation of short- and medium-term memories, which allow the system to better manage object identities across frames and sequences. Our novel memory aggregation method also contributes to improving the model's performance by effectively consolidating different memory durations. Moreover, STMMOT showed excellent generalization and adaptability when dealing with different feature extraction backbone networks, proving its flexibility and wide applicability. Our tests on the MOT17 and MOT20 test sets confirmed the model's superior tracking accuracy, leading to a notable reduction in identity switches. While the results of STMMOT are encouraging, future work could explore more advanced fusion techniques or leverage more sophisticated transformer architectures to further enhance the model's tracking capabilities.

## 8.2.   FUTURE WORK

There are multiple future directions that our research can pursue to further strengthen the effectiveness of the proposed multi-object tracking (MOT) model. One potential area of exploration is the optimization of the attention mechanism. Our proposed model employs an attention mechanism to manage the interactions between multiple objects, especially in crowded scenes. However, this mechanism could potentially be further refined to enhance its accuracy and efficiency in dealing with such complex scenarios. Future work may involve implementing various optimization strategies, perhaps by incorporating additional context information or adjusting the attention computation mechanism to better cope with densely populated scenes. Another promising line of inquiry lies in testing our proposed model in different domains. While the model has demonstrated promising results in the sports analytics context, it is critical to examine its applicability and generalization ability in various other scenarios. For instance, the model could be tested in domains such as surveillance, where robust tracking of multiple objects is crucial for maintaining security. Similarly, autonomous driving and robotics present challenging and dynamic environments in which robust multi-object tracking can play a critical role. The model's performance in these different domains would offer valuable insights into its versatility and adaptability. Moreover, scalability remains an essential aspect to be considered in the future development of our MOT model. As the number of track objects increases, maintaining the consistency of identities (avoiding ID switches) becomes a more challenging task. Hence, future research efforts should focus on improving the model's scalability, ensuring that its performance remains stable even when faced with a higher number of tracked objects. Possible strategies could involve improving the memory management mechanisms or implementing more advanced data association techniques.

Finally, future research could also look into combining our proposed framework with other existing methods for MOT. Even though our model has shown significant improvements over the existing methods, there may still be useful techniques or components in those methods that could further improve the performance and robustness of our model when integrated properly. The fusion of different methods might lead to the development of a more powerful and versatile MOT system that can efficiently handle various challenges in different scenarios